%% file: main.tex
\documentclass[runningheads]{llncs}
\usepackage[T1]{fontenc}
\usepackage{graphicx,verbatim}
\usepackage{amsfonts,amssymb,amsmath}
\usepackage{xspace}
\usepackage{bm}
\usepackage{booktabs}
\usepackage{multirow}
\usepackage[table,xcdraw]{xcolor}
\usepackage{arydshln}
\usepackage{colortbl}
\usepackage[colorlinks]{hyperref}
\usepackage{cleveref}

\crefname{equation}{Eqn.}{Eqns.}
\Crefname{equation}{Eqn.}{Eqns.}

\newcommand{\seg}{segmentation\xspace}
\newcommand{\segs}{segmentations\xspace}
\newcommand{\semisup}{semi-supervised\xspace}

\newcommand{\qual}{quality\xspace}
\newcommand{\segq}{segmentation quality\xspace}
\newcommand{\gt}{ground truth\xspace}
\newcommand{\pl}{pseudolabel\xspace}
\newcommand{\pls}{pseudolabels\xspace}

\newcommand{\dl}{\mathcal{D}_L}
\newcommand{\du}{\mathcal{D}_U}
\newcommand{\dq}{\mathcal{D}_Q}
\newcommand{\mq}{g_\phi}
\newcommand{\segparams}{\theta}
\newcommand{\segmodel}{f_\segparams}
\newcommand{\corruptionfn}{h}
\newcommand{\lsup}{\mathcal{L}_{\mathrm{sup}}}
\newcommand{\lqar}{\mathcal{L}_{\mathrm{qar}}}
\newcommand{\lqw}{\mathcal{L}_{\mathrm{qw}}}

\newcommand{\ie}{{i.e.,}\xspace}
\newcommand{\eg}{{e.g.,}\xspace}

\newcommand{\rev}[1]{{#1}}

\usepackage{url}
\newcommand{\ghrepourl}{\url{https://github.com/sfu-mial/QG-SSL}}

\definecolor{posLow}{HTML}{D0EED1} 
\definecolor{posMidLow}{HTML}{B7E1B9}
\definecolor{posMidHigh}{HTML}{A5D6A7}
\definecolor{posHigh}{HTML}{81C784} 
\definecolor{negLow}{HTML}{FFEBEE} 

\definecolor{lightgreyHLcolor}{HTML}{F0F0F0}
\definecolor{mediumgreyHeading}{HTML}{D0D0D0}

\newcommand{\greyCellColor}{\cellcolor{mediumgreyHeading}}

\newcommand{\posLowBox}[1]{\colorbox{posLow}{\strut #1}}
\newcommand{\posMidLowBox}[1]{\colorbox{posMidLow}{\strut #1}}
\newcommand{\posMidHighBox}[1]{\colorbox{posMidHigh}{\strut #1}}
\newcommand{\posHighBox}[1]{\colorbox{posHigh}{\strut #1}}

\newcommand{\negLowBox}[1]{\colorbox{negLow}{\strut #1}}

\newcommand{\lightgreyHLBox}[1]{\colorbox{lightgreyHLcolor}{#1}}

\newcommand{\cellvrule}[1]{%
  \multicolumn{1}{c@{\hspace{0.4em}\smash{\rule[-0.7ex]{0.5pt}{3.0ex}}\hspace{0.4em}}}{#1}%
}

\begin{document}
\title{Quality-Guided Semi-Supervised Learning for Medical Image Segmentation}
%
\author{Kumar Abhishek\orcidID{0000-0002-7341-9617} \and
Ghassan Hamarneh\orcidID{0000-0001-5040-7448}}
%
\authorrunning{K. Abhishek et al.}
%
\institute{School of Computing Science, Simon Fraser University, Canada
\email{\{kabhishe,hamarneh\}@sfu.ca}}



\maketitle              
\begin{abstract}

\input{sections/abstract}

\end{abstract}

\input{sections/introduction}

\input{sections/method}

\input{sections/results}

\input{sections/conclusion}

\begin{credits}
\subsubsection{\ackname} 
\rev{
The authors thank Darren Sutton for initial discussions and acknowledge computational support from NVIDIA Corporation and the Digital Research Alliance of Canada. Partial funding for this project was provided by the Natural Sciences and Engineering Research Council of Canada (NSERC RGPIN-2020-06752).
}

\subsubsection{\discintname}
\rev{
The authors have no competing interests to declare.
}
\end{credits}

\bibliographystyle{splncs04}
\bibliography{references}

\end{document}

%% file: sections/abstract.tex
Training accurate medical image segmentation models requires large amounts of densely annotated data, which is costly and time-consuming to obtain. Semi-supervised learning (SSL) alleviates this by learning from both abundant unlabeled data and limited labeled data. However, most modern SSL methods rely on pseudolabels for unlabeled data, and typically assess their reliability through model confidence or uncertainty, measures that are self-referential and lack explicit grounding in segmentation quality. 
Instead, we propose a \textit{quality}-guided SSL framework that trains a dedicated network to estimate segmentation quality from image-mask pairs. The predictor is trained on variable-quality masks generated through synthetic corruptions augmented with imperfect outputs from partially trained segmentation models, capturing realistic error patterns encountered during training. We integrate the quality predictor into SSL through two complementary mechanisms: a quality-aware regularization loss and a quality-based pseudolabel sample reweighting scheme. We show that our method serves as a drop-in enhancement to existing SSL frameworks. Extensive experiments across five datasets and multiple architectures demonstrate consistent improvements over competing SSL methods, advancing the state-of-the-art in semi-supervised medical image segmentation.

\keywords{Image segmentation \and Segmentation quality  \and Semi-supervised learning.}

%% file: sections/introduction.tex
\section{Introduction}

Accurate \seg of medical images is fundamental to clinical workflows, yet dense pixelwise annotations, necessary for training deep learning-based \seg models, remain costly and scarce~\cite{litjens2017survey,tajbakhsh2020embracing,asgari2021deep}.
Semi-supervised learning (SSL) addresses this annotation scarcity by leveraging abundant unlabeled data alongside limited labels, and has become a dominant paradigm for label-efficient medical image \seg. 

Most existing SSL approaches differ in how they leverage unlabeled data, falling into three broad categories: (i)~{consistency regularization}, enforcing prediction invariance under perturbations, notably mean teacher (MT)~\cite{tarvainen2017mean}, its uncertainty-aware extension (UA-MT)~\cite{yu2019uncertainty}, and interpolation consistency training (ICT)~\cite{verma2022interpolation}; (ii)~{\pl methods}, using confident predictions as surrogate supervision \cite{lee2013pseudo,sohn2020fixmatch}, extended by cross-pseudo supervision (CPS)~\cite{chen2021semi};
and (iii)~{contrastive learning}, leveraging representation-level objectives on unlabeled data~\cite{chaitanya2023local}. Across these methods, unlabeled samples are either treated uniformly regardless of prediction \qual (MT, ICT), or filtered using model-derived confidence as a proxy for reliability (UA-MT, FixMatch~\cite{sohn2020fixmatch}). 
Although calibration techniques can reduce overconfidence~\cite{guo2017calibration}, medical \seg networks remain poorly calibrated in practice~\cite{mehrtash2020confidence}. More fundamentally, even perfectly calibrated confidence is {self-referential}, since it reflects the model's belief about its own prediction, and cannot catch systematic errors arising from the same representations that produced them. 
We argue that an independent assessment of \segq is a better alternative to model certainty for guiding SSL training. 

Predicting \segq without \gt has been studied for clinical \qual control. Early work predicted Dice scores from hand-crafted features~\cite{kohlberger2012evaluating} and reverse classifiers to estimate \qual~\cite{valindria2017reverse}. More recent approaches directly regress \qual metrics from image-\seg pairs~\cite{robinson2018real,devries2018leveraging,qiu2023qcresunet}, and large-scale models now offer general-purpose \qual prediction across diverse anatomies~\cite{senbi2024towards}. However, this entire line of work treats \qual prediction as an end goal, filtering unreliable \segs post-hoc. No prior work has leveraged learned \qual prediction to guide \semisup training itself.

We bridge these two directions by training a \qual predictor that estimates \segq from image-mask pairs, then using it to provide learning signal for unlabeled data in SSL. Unlike confidence from a single forward pass, 
predicted quality provides a complementary signal by comparing mask structure against image evidence, independently of the \seg network's own representations.
While related to sample reweighting for noisy labels~\cite{ren2018learning,shu2019meta}, 
our predictor directly estimates \seg accuracy rather than inferring importance from per-sample loss values.
However, a quality predictor trained on variable quality masks from labeled data must generalize to real network predictions of unlabeled data.

\input{figs/overview} 

To provide a quality-based guidance for medical image \seg in a semi-supervised setting, we contribute:
(1)~the first framework-agnostic method to leverage learned \segq prediction for guiding SSL; (2)~two complementary mechanisms for integrating quality predictions into SSL training: a differentiable \qual regularizer and a \pl reweighting scheme, applicable as drop-in enhancements to existing SSL methods; (3)~a mask corruption strategy incorporating partially-trained model predictions that exhibit characteristic neural network errors~\cite{byrne2021persistent,kirillov2020pointrend} to bridge the distribution gap between synthetic and real network errors; and (4)~we perform comprehensive experiments across five cross-dataset pairs spanning dermatology and colonoscopy, five SSL paradigms, and multiple model architectures, yielding consistent improvements over state-of-the-art baselines. Our code is publicly available at \ghrepourl.

%% file: figs/overview.tex
\begin{figure*}[ht!]
    \centering
        \includegraphics[width=0.9\textwidth]{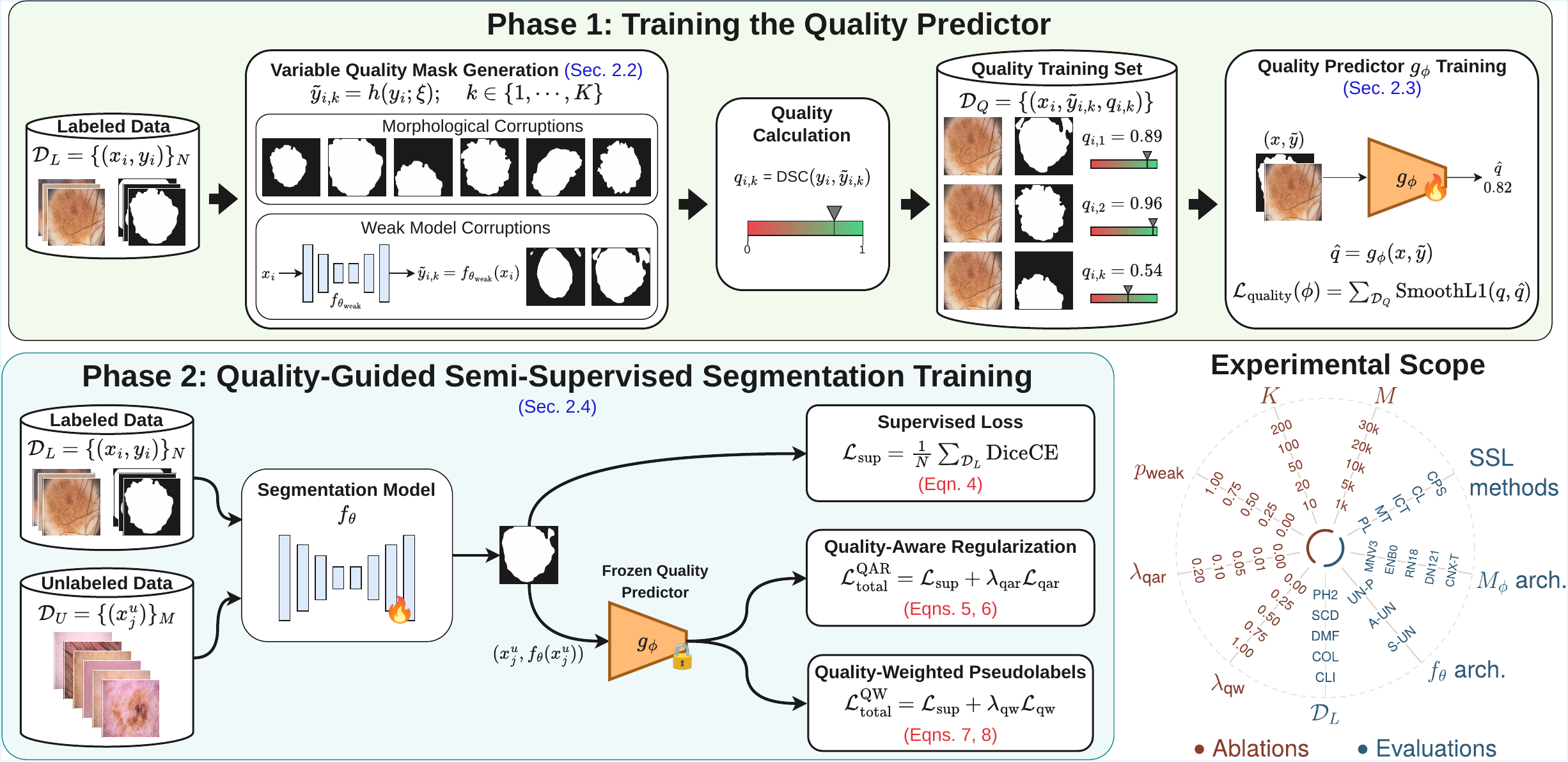}
        \caption{
        An overview of the proposed \qual-guided \semisup \seg methods, along with the scope of experiments present in this paper.
        }
        \label{fig:overview}
\end{figure*}

%% file: sections/method.tex
\section{Method}
\label{sec:method}

Our proposed approach has two phases: {(Phase 1)}~training a \qual predictor $\mq$ to estimate \segq from image-mask pairs, and {(Phase 2)}~using the frozen $\mq$ to guide \semisup \seg training. Fig.~\ref{fig:overview} provides an overview and the scope of experiments presented in this paper.

\subsection{Problem Definition}
\label{subsec:problem}

Let $\dl = \{(x_i, y_i)\}_{i=1}^{N}$ denote a small labeled set, where $x_i \in \mathbb{R}^{H \times W \times C}$ is an image and $y_i \in \{0,1\}^{H \times W}$ is its \gt binary \seg mask, and let $\du = \{x^u_j\}_{j=1}^{M}$ ($M \gg N$) be a larger unlabeled set. Our goal is to train a \seg network $\segmodel$ that leverages both $\dl$ and $\du$.
In our experiments, $\dl$ and $\du$ come from related but distinct sources (\eg PH2~\cite{mendonca2013ph} and ISIC2020~\cite{rotemberg2021patient}), a challenging setting that better reflects clinical practice. We do not assume matched distributions between labeled and unlabeled data.

\subsection{Variable Quality Mask Generation}
\label{subsec:vqm}

To train $\mq$, we require images paired with arbitrary masks, where each mask has a corresponding \qual score. We construct a synthetic dataset $\dq$ from $\dl$ by generating, for each $(x_i, y_i) \in \dl$, a set of $K$ degraded masks using a stochastic corruption function $\corruptionfn$:
\begin{align}
\label{eqn:vqm}
    \tilde{y}_{i,k} = \corruptionfn(y_i;\, \xi_k), \qquad q_{i,k} = \mathrm{DSC}(y_i,\, \tilde{y}_{i,k}), \qquad k \in \{1, \cdots K\},
\end{align}

\noindent where $\xi_k$ represents random perturbation parameters, and $q_{i,k} \in [0, 1]$ is the Dice score (DSC) of each corrupted mask. We sample two types of degradations randomly. First (Type 1), we use random morphological operations (erosion/dilation with different kernel sizes), translations, elastic deformations, additive noise, and boundary perturbations. However, morphological corruptions alone may not capture error patterns produced by real neural networks during \semisup training. To bridge this distribution gap, (Type 2) we augment our corruption strategy with predictions from {partially trained} (weak) \seg models $f_{\segparams_\mathrm{weak}}$. We train a U-Net~\cite{ronneberger2015u} on $\dl$ from random initialization and collect checkpoints at early epochs (epochs 1, 3, 5, 10, 15, 20).
These weak models produce \seg predictions exhibiting characteristic early training failure patterns~\cite{kirillov2020pointrend,byrne2021persistent}.
Previous work on learned \qual prediction has also noted that limited corruption diversity restricts the predictor's sensitivity to fine-grained \qual differences~\cite{robinson2018real}, further motivating the inclusion of real network outputs of imperfect \seg models. Therefore, when training $\mq$, we sample with probability $p_\mathrm{weak}$ from these weak-model predictions (Type 2) and $1 - p_\mathrm{weak}$ from Type 1 corruptions. The resulting dataset $\dq = \{(x_i, \tilde{y}_{i,k}, q_{i,k})\}$ contains image-mask-\qual triplets spanning the full range of Dice scores. 
We explore various settings of $p_\mathrm{weak}$ and $K$.

\subsection{Segmentation Quality Predictor}
\label{subsec:quality_predictor}

The \qual predictor $\mq$ takes 
an image-mask pair, outputs a scalar \qual estimate (Eqn.~\ref{eqn:m_q_prediction}) and is trained to minimize a regression loss \rev{$\ell_{\mathrm{reg}}$} (Eqn.~\ref{eqn:l_quality}):
\begin{align}
    \hat{q} &= \mq(x , \tilde{y}), \label{eqn:m_q_prediction} \\
    \mathcal{L}_\mathrm{quality}(\phi) &= \sum_{(x,\, \tilde{y},\, q)\, \in\, \dq} \ell_{\mathrm{reg}}\!\left(\mq(x , \tilde{y}),\; q\right) \label{eqn:l_quality}.
\end{align}

\noindent A key property distinguishing \qual prediction from proxy signals such as model confidence is that it is designed to be contextually grounded: to assess whether $\tilde{y}$ is a good \seg of $x$, $\mq$ must compare mask structure against visual evidence in the image, rather than relying on the mask or the model's internal state alone. Once trained, $\mq$ is frozen and acts as a differentiable \qual assessment function for any image-mask pair without requiring \gt.

\subsection{Quality-Guided Semi-Supervised Training}
\label{subsec:ssl_training}

We now train $\segmodel$ using both $\dl$ and $\du$. 
For all labeled samples, we minimize:
\begin{align}
\label{eqn:seg_loss}
    \lsup(\segparams) = \frac{1}{N} \sum_{i=1}^{N} \ell_{\mathrm{seg}}\!\left(\segmodel(x_i),\; y_i\right).
\end{align}

\noindent For unlabeled data, we propose two alternative mechanisms for leveraging the frozen $\mq$, differing in 
whether the \seg loss \rev{$\ell_{\mathrm{seg}}$} gradients propagate through $\mq$ {(QAR)} or $\mq$ serves only to compute per-sample weights {(PL-QW)}.

\medskip
\noindent\textbf{A: {Q}uality-{A}ware {R}egularization (QAR):}
For each unlabeled sample $x^u_j$, the soft prediction $\segmodel(x^u_j)$ is passed into $\mq$. No explicit \pls are generated; instead, gradients of the loss $\lqar$ propagate from the scalar \qual output back through the predicted mask into \seg model parameters $\segparams$, encouraging $\segmodel$ to produce \segs that $\mq$ judges as high \qual:
\begin{align}
\label{eqn:l_reg}
    \lqar(\segparams) = \frac{1}{M} \sum_{j=1}^{M} \left(1 - \mq\!\left(x^u_j , \segmodel(x^u_j)\right)\right).
\end{align}

\noindent The complete objective is a weighted sum of the two losses:
\begin{align}
\label{eqn:total_reg}
    \mathcal{L}_\mathrm{total}^\mathrm{QAR} = \lsup + \lambda_\mathrm{qar}\, \lqar.
\end{align}

\medskip
\noindent\textbf{B: {Q}uality-{W}eighted Pseudolabels (PL-QW):}
Given \pls $\hat{y}^u_j$ for unlabeled samples $x^u_j \in \du$, we weight the per-sample loss by predicted \pl \qual:
\begin{align}
\label{eqn:l_qw}
    \lqw(\segparams) = \frac{1}{M} \sum_{j=1}^{M} w_j \cdot \ell_{\mathrm{seg}}\!\left(\segmodel(x^u_j),\; \hat{y}^u_j\right),
\end{align}
where $w_j = \mq(x^u_j , \hat{y}^u_j)$ is computed with $\mq$ frozen and detached from the computational graph. Unlike QAR, no gradients flow through $\mq$; it serves purely as a sample weighting function that upweights high-\qual \pls and downweights unreliable ones. The complete objective is:
\begin{align}
\label{eqn:total_qw}
    \mathcal{L}_\mathrm{total}^\mathrm{QW} = \lsup + \lambda_\mathrm{qw}\, \lqw.
\end{align}

\noindent A key property of this formulation is its orthogonality to the choice of \semisup method: any approach generating \pls $\hat{y}^u_j$ can be augmented by weighting per-sample losses with $w_j$, without requiring architectural changes.

\input{tables/bigtable-hltext}

%% file: tables/bigtable-hltext.tex
\begin{table}[!ht]
\centering
\caption{Quantitative results (DSC and IoU; mean{\tiny \emph{std.err.}}) 
for all SSL baselines, their quality-weighted versions, and QAR.
Using a quality predictor consistently improves \seg performance across 5 datasets and 3 \seg model architectures.}
\label{tab:results}
\resizebox{\textwidth}{!}{%
\setlength{\tabcolsep}{0.5em}
\def\arraystretch{1.15}
\setlength{\fboxsep}{2pt}
\begin{tabular}{@{}ccc|c|ccc|ccc|cc|cc|cc|c@{}}
\toprule

&  &  &  & \multicolumn{3}{c}{\emph{Pseudolabels}} & \multicolumn{3}{c}{\emph{Student-Teacher Models}} & \multicolumn{2}{c}{\emph{Interp. Consistency}} & \multicolumn{2}{c}{\emph{Contrastive}} & \multicolumn{2}{c}{\emph{Cross Pseudo Sup.}} &  \\
\multirow{-2}{*}{\textbf{$\dl$}} & \multirow{-2}{*}{\textbf{\begin{tabular}[c]{@{}c@{}}$\segmodel$\\ Arch.\end{tabular}}} & \multirow{-2}{*}{\textbf{}} & \multirow{-2}{*}{\textbf{SUP}} & \textbf{\begin{tabular}[c]{@{}c@{}}PL-T\\\cite{lee2013pseudo}\end{tabular}} & \textbf{\begin{tabular}[c]{@{}c@{}}PL-C\\\cite{sohn2020fixmatch}\end{tabular}} & \textbf{\begin{tabular}[c]{@{}c@{}}PL-QW\\ (Ours)\end{tabular}} & \textbf{\begin{tabular}[c]{@{}c@{}}MT\\\cite{tarvainen2017mean}\end{tabular}} & \textbf{\begin{tabular}[c]{@{}c@{}}UA-MT\\\cite{yu2019uncertainty}\end{tabular}} & \textbf{\begin{tabular}[c]{@{}c@{}}MT-QW\\ (Ours)\end{tabular}} & \textbf{\begin{tabular}[c]{@{}c@{}}ICT\\\cite{verma2022interpolation}\end{tabular}} & \textbf{\begin{tabular}[c]{@{}c@{}}ICT-QW\\ (Ours)\end{tabular}} & \textbf{\begin{tabular}[c]{@{}c@{}}CL\\\cite{chaitanya2023local}\end{tabular}} & \textbf{\begin{tabular}[c]{@{}c@{}}CL-QW\\ (Ours)\end{tabular}} & \textbf{\begin{tabular}[c]{@{}c@{}}CPS\\\cite{chen2021semi}\end{tabular}} & \textbf{\begin{tabular}[c]{@{}c@{}}CPS-QW\\ (Ours)\end{tabular}} & \multirow{-2}{*}{\textbf{\begin{tabular}[c]{@{}c@{}}QAR\\ (Ours)\end{tabular}}} \\
\midrule
&  & DSC & 92.93\tiny{\emph{0.52}} & 93.21\tiny{\emph{0.88}} & 93.92\tiny{\emph{1.10}} & \posHighBox{95.46\tiny{\emph{0.45}}} & 93.53\tiny{\emph{1.62}} & 94.33\tiny{\emph{0.62}} & \posHighBox{95.32\tiny{\emph{0.43}}} & 93.85\tiny{\emph{0.82}} & \posMidHighBox{95.18\tiny{\emph{0.46}}} & 92.37\tiny{\emph{1.82}} & \posHighBox{94.48\tiny{\emph{0.56}}} & 94.20\tiny{\emph{0.71}} & \posMidLowBox{95.19\tiny{\emph{0.52}}} & \posHighBox{95.48\tiny{\emph{0.48}}} \\
& \multirow{-2}{*}{UN-P} & IoU & 86.96\tiny{\emph{0.89}} & 87.75\tiny{\emph{1.41}} & 89.19\tiny{\emph{1.58}} & \posHighBox{91.45\tiny{\emph{0.79}}} & 89.04\tiny{\emph{2.00}} & 89.51\tiny{\emph{1.04}} & \posHighBox{91.18\tiny{\emph{0.76}}} & 88.81\tiny{\emph{1.30}} & \posHighBox{90.95\tiny{\emph{0.81}}} & 87.23\tiny{\emph{2.12}} & \posHighBox{89.75\tiny{\emph{0.97}}} & 89.46\tiny{\emph{1.16}} & \posHighBox{91.01\tiny{\emph{0.91}}} & \posHighBox{91.50\tiny{\emph{0.84}}} \\
\cdashline{2-17}
&  & DSC & 91.73\tiny{\emph{0.69}} & 92.90\tiny{\emph{0.67}} & 93.34\tiny{\emph{0.81}} & \posHighBox{95.40\tiny{\emph{0.45}}} & 93.18\tiny{\emph{1.60}} & 93.70\tiny{\emph{0.64}} & \posHighBox{95.38\tiny{\emph{0.44}}} & 94.16\tiny{\emph{0.58}} & \posMidHighBox{95.17\tiny{\emph{0.49}}} & 92.91\tiny{\emph{1.88}} & \posHighBox{95.09\tiny{\emph{0.48}}} & 93.91\tiny{\emph{0.70}} & \posMidHighBox{94.99\tiny{\emph{0.52}}} & \posHighBox{95.29\tiny{\emph{0.47}}} \\
& \multirow{-2}{*}{A-UN} & IoU & 85.01\tiny{\emph{1.12}} & 87.17\tiny{\emph{1.12}} & 87.92\tiny{\emph{1.32}} & \posHighBox{91.33\tiny{\emph{0.80}}} & 88.40\tiny{\emph{1.98}} & 88.41\tiny{\emph{1.08}} & \posHighBox{91.31\tiny{\emph{0.78}}} & 89.18\tiny{\emph{0.99}} & \posHighBox{90.95\tiny{\emph{0.86}}} & 88.28\tiny{\emph{2.20}} & \posHighBox{90.80\tiny{\emph{0.84}}} & 88.83\tiny{\emph{1.16}} & \posHighBox{90.64\tiny{\emph{0.89}}} & \posHighBox{91.15\tiny{\emph{0.82}}} \\
\cdashline{2-17}
&  & DSC & 93.02\tiny{\emph{0.55}} & 94.50\tiny{\emph{0.75}} & 94.86\tiny{\emph{0.50}} & \posMidHighBox{95.57\tiny{\emph{0.41}}} & 94.90\tiny{\emph{0.47}} & 94.87\tiny{\emph{0.46}} & \posLowBox{95.36\tiny{\emph{0.44}}} & 94.69\tiny{\emph{0.50}} & \posMidLowBox{95.26\tiny{\emph{0.40}}} & 93.63\tiny{\emph{1.65}} & \posHighBox{95.70\tiny{\emph{0.39}}} & 94.32\tiny{\emph{0.49}} & \posMidHighBox{95.45\tiny{\emph{0.44}}} & \posHighBox{95.58\tiny{\emph{0.42}}} \\
\multirow{-6}{*}{PH2} & \multirow{-2}{*}{S-UN} & IoU & 87.14\tiny{\emph{0.94}} & 90.05\tiny{\emph{1.22}} & 90.39\tiny{\emph{0.87}} & \posHighBox{91.63\tiny{\emph{0.74}}} & 90.63\tiny{\emph{0.84}} & 90.30\tiny{\emph{0.81}} & \posMidLowBox{91.27\tiny{\emph{0.78}}} & 90.08\tiny{\emph{0.89}} & \posMidLowBox{91.05\tiny{\emph{0.72}}} & 89.21\tiny{\emph{1.95}} & \posHighBox{91.86\tiny{\emph{0.70}}} & 89.42\tiny{\emph{0.85}} & \posHighBox{91.42\tiny{\emph{0.78}}} & \posHighBox{91.66\tiny{\emph{0.75}}} \\

\midrule

&  & DSC & 89.88\tiny{\emph{0.95}} & 91.39\tiny{\emph{1.84}} & 91.87\tiny{\emph{1.16}} & \posHighBox{93.18\tiny{\emph{0.73}}} & 91.46\tiny{\emph{1.19}} & 92.20\tiny{\emph{0.93}} & \posMidHighBox{92.91\tiny{\emph{0.88}}} & 91.86\tiny{\emph{1.31}} & \posHighBox{93.61\tiny{\emph{0.50}}} & 92.37\tiny{\emph{1.13}} & \posLowBox{92.67\tiny{\emph{1.02}}} & 90.06\tiny{\emph{0.90}} & \posHighBox{93.26\tiny{\emph{0.75}}} & \posHighBox{93.24\tiny{\emph{0.69}}} \\
& \multirow{-2}{*}{UN-P} & IoU & 82.15\tiny{\emph{1.46}} & 85.70\tiny{\emph{2.21}} & 85.73\tiny{\emph{1.71}} & \posHighBox{87.57\tiny{\emph{1.18}}} & 85.08\tiny{\emph{1.76}} & 86.05\tiny{\emph{1.45}} & \posHighBox{87.22\tiny{\emph{1.35}}} & 85.87\tiny{\emph{1.84}} & \posHighBox{88.16\tiny{\emph{0.87}}} & 86.52\tiny{\emph{1.59}} & \posLowBox{86.96\tiny{\emph{1.54}}} & 82.39\tiny{\emph{1.40}} & \posHighBox{87.72\tiny{\emph{1.21}}} & \posHighBox{87.63\tiny{\emph{1.11}}} \\
\cdashline{2-17}
&  & DSC & 90.89\tiny{\emph{0.89}} & 91.71\tiny{\emph{1.68}} & 91.83\tiny{\emph{1.02}} & \posMidHighBox{93.15\tiny{\emph{0.76}}} & 92.23\tiny{\emph{1.22}} & 92.40\tiny{\emph{0.90}} & \posMidLowBox{93.06\tiny{\emph{0.81}}} & 92.34\tiny{\emph{0.98}} & \posMidLowBox{93.04\tiny{\emph{0.73}}} & 92.42\tiny{\emph{0.82}} & \posLowBox{92.64\tiny{\emph{0.88}}} & 92.64\tiny{\emph{0.72}} & \posMidLowBox{93.62\tiny{\emph{0.65}}} & \posHighBox{93.11\tiny{\emph{0.72}}} \\
& \multirow{-2}{*}{A-UN} & IoU & 83.78\tiny{\emph{1.39}} & 85.99\tiny{\emph{2.02}} & 85.50\tiny{\emph{1.54}} & \posHighBox{87.54\tiny{\emph{1.21}}} & 86.41\tiny{\emph{1.75}} & 86.35\tiny{\emph{1.39}} & \posMidHighBox{87.42\tiny{\emph{1.27}}} & 86.34\tiny{\emph{1.51}} & \posMidLowBox{87.31\tiny{\emph{1.17}}} & 86.33\tiny{\emph{1.30}} & \posLowBox{86.76\tiny{\emph{1.38}}} & 86.61\tiny{\emph{1.16}} & \posHighBox{88.26\tiny{\emph{1.04}}} & \posHighBox{87.44\tiny{\emph{1.17}}} \\
\cdashline{2-17}
&  & DSC & 91.54\tiny{\emph{0.58}} & 92.60\tiny{\emph{0.87}} & 92.69\tiny{\emph{0.70}} & \posMidHighBox{93.84\tiny{\emph{0.47}}} & 92.71\tiny{\emph{0.53}} & 93.30\tiny{\emph{0.55}} & \posMidLowBox{93.69\tiny{\emph{0.46}}} & 92.94\tiny{\emph{0.45}} & \posMidLowBox{93.64\tiny{\emph{0.52}}} & 93.07\tiny{\emph{0.65}} & \posMidLowBox{93.71\tiny{\emph{0.47}}} & 93.30\tiny{\emph{0.54}} & \posLowBox{93.72\tiny{\emph{0.49}}} & \posHighBox{93.77\tiny{\emph{0.46}}} \\
\multirow{-6}{*}{SCD} & \multirow{-2}{*}{S-UN} & IoU & 84.62\tiny{\emph{0.96}} & 86.70\tiny{\emph{1.36}} & 86.60\tiny{\emph{1.15}} & \posHighBox{88.55\tiny{\emph{0.81}}} & 86.59\tiny{\emph{0.89}} & 87.80\tiny{\emph{0.93}} & \posHighBox{88.27\tiny{\emph{0.80}}} & 86.90\tiny{\emph{0.78}} & \posMidHighBox{88.22\tiny{\emph{0.89}}} & 87.31\tiny{\emph{1.07}} & \posMidHighBox{88.31\tiny{\emph{0.81}}} & 87.70\tiny{\emph{0.91}} & \posMidLowBox{88.35\tiny{\emph{0.84}}} & \posHighBox{88.41\tiny{\emph{0.79}}} \\

\midrule

&  & DSC & 90.00\tiny{\emph{0.52}} & 90.10\tiny{\emph{0.51}} & 90.30\tiny{\emph{0.50}} & \posMidHighBox{91.34\tiny{\emph{0.44}}} & 90.53\tiny{\emph{0.50}} & 90.61\tiny{\emph{0.47}} & \posMidLowBox{91.24\tiny{\emph{0.44}}} & 90.91\tiny{\emph{0.47}} & \posLowBox{91.35\tiny{\emph{0.43}}} & 91.04\tiny{\emph{0.46}} & \negLowBox{91.01\tiny{\emph{0.45}}} & 90.44\tiny{\emph{0.49}} & \posMidLowBox{91.22\tiny{\emph{0.43}}} & \posMidHighBox{91.34\tiny{\emph{0.43}}} \\
& \multirow{-2}{*}{UN-P} & IoU & 82.87\tiny{\emph{0.73}} & 82.89\tiny{\emph{0.75}} & 83.31\tiny{\emph{0.74}} & \posHighBox{84.76\tiny{\emph{0.67}}} & 83.56\tiny{\emph{0.74}} & 83.62\tiny{\emph{0.71}} & \posMidHighBox{84.58\tiny{\emph{0.67}}} & 84.10\tiny{\emph{0.69}} & \posMidLowBox{84.75\tiny{\emph{0.65}}} & 84.29\tiny{\emph{0.68}} & \negLowBox{84.24\tiny{\emph{0.68}}} & 83.36\tiny{\emph{0.71}} & \posMidHighBox{84.54\tiny{\emph{0.66}}} & \posHighBox{84.74\tiny{\emph{0.66}}} \\
\cdashline{2-17}
&  & DSC & 90.06\tiny{\emph{0.47}} & 90.35\tiny{\emph{0.46}} & 90.54\tiny{\emph{0.45}} & \posMidLowBox{91.31\tiny{\emph{0.42}}} & 90.60\tiny{\emph{0.44}} & 90.67\tiny{\emph{0.46}} & \posMidLowBox{91.27\tiny{\emph{0.42}}} & 90.53\tiny{\emph{0.50}} & \posMidLowBox{91.21\tiny{\emph{0.43}}} & 90.38\tiny{\emph{0.48}} & \posMidHighBox{91.46\tiny{\emph{0.42}}} & 90.56\tiny{\emph{0.46}} & \posMidLowBox{91.25\tiny{\emph{0.43}}} & \posMidHighBox{91.37\tiny{\emph{0.44}}} \\
& \multirow{-2}{*}{A-UN} & IoU & 82.68\tiny{\emph{0.69}} & 83.15\tiny{\emph{0.69}} & 83.44\tiny{\emph{0.68}} & \posHighBox{84.65\tiny{\emph{0.65}}} & 83.68\tiny{\emph{0.68}} & 83.69\tiny{\emph{0.70}} & \posMidLowBox{84.59\tiny{\emph{0.64}}} & 83.57\tiny{\emph{0.74}} & \posMidLowBox{84.50\tiny{\emph{0.66}}} & 83.26\tiny{\emph{0.71}} & \posHighBox{84.92\tiny{\emph{0.65}}} & 83.49\tiny{\emph{0.69}} & \posMidHighBox{84.58\tiny{\emph{0.66}}} & \posHighBox{84.81\tiny{\emph{0.67}}} \\
\cdashline{2-17}
&  & DSC & 90.47\tiny{\emph{0.45}} & 91.17\tiny{\emph{0.45}} & 91.21\tiny{\emph{0.42}} & \posMidLowBox{91.94\tiny{\emph{0.41}}} & 91.24\tiny{\emph{0.43}} & 91.17\tiny{\emph{0.43}} & \posLowBox{91.67\tiny{\emph{0.42}}} & 91.21\tiny{\emph{0.45}} & \posMidLowBox{91.84\tiny{\emph{0.41}}} & 91.22\tiny{\emph{0.45}} & \posLowBox{91.67\tiny{\emph{0.42}}} & 91.60\tiny{\emph{0.40}} & \posLowBox{92.08\tiny{\emph{0.40}}} & \posMidHighBox{91.89\tiny{\emph{0.40}}} \\
\multirow{-6}{*}{DMF} & \multirow{-2}{*}{S-UN} & IoU & 83.30\tiny{\emph{0.67}} & 84.50\tiny{\emph{0.68}} & 84.49\tiny{\emph{0.65}} & \posMidHighBox{85.72\tiny{\emph{0.64}}} & 84.57\tiny{\emph{0.65}} & 84.43\tiny{\emph{0.65}} & \posMidLowBox{85.28\tiny{\emph{0.65}}} & 84.56\tiny{\emph{0.68}} & \posMidLowBox{85.54\tiny{\emph{0.64}}} & 84.59\tiny{\emph{0.68}} & \posMidLowBox{85.27\tiny{\emph{0.65}}} & 85.23\tiny{\emph{0.62}} & \posMidLowBox{85.92\tiny{\emph{0.62}}} & \posHighBox{85.59\tiny{\emph{0.62}}} \\

\midrule

&  & DSC & 89.17\tiny{\emph{1.82}} & 89.59\tiny{\emph{1.81}} & 88.89\tiny{\emph{1.49}} & \posHighBox{91.73\tiny{\emph{1.19}}} & 90.56\tiny{\emph{1.86}} & 90.10\tiny{\emph{1.85}} & \posMidHighBox{91.64\tiny{\emph{1.21}}} & 90.72\tiny{\emph{1.45}} & \posLowBox{91.03\tiny{\emph{1.59}}} & 89.57\tiny{\emph{1.83}} & \posMidHighBox{90.73\tiny{\emph{1.41}}} & 89.60\tiny{\emph{1.71}} & \posMidHighBox{90.62\tiny{\emph{1.69}}} & \posHighBox{92.01\tiny{\emph{1.02}}} \\
& \multirow{-2}{*}{UN-P} & IoU & 82.93\tiny{\emph{2.05}} & 83.53\tiny{\emph{1.95}} & 81.99\tiny{\emph{1.96}} & \posHighBox{86.01\tiny{\emph{1.56}}} & 85.23\tiny{\emph{1.97}} & 84.40\tiny{\emph{1.94}} & \posMidLowBox{85.93\tiny{\emph{1.60}}} & 84.81\tiny{\emph{1.81}} & \posMidLowBox{85.46\tiny{\emph{1.78}}} & 83.69\tiny{\emph{2.08}} & \posMidHighBox{84.79\tiny{\emph{1.82}}} & 83.45\tiny{\emph{1.99}} & \posHighBox{85.04\tiny{\emph{1.90}}} & \posHighBox{86.24\tiny{\emph{1.45}}} \\
\cdashline{2-17}
&  & DSC & 89.25\tiny{\emph{1.94}} & 88.38\tiny{\emph{1.91}} & 89.19\tiny{\emph{1.88}} & \posHighBox{91.37\tiny{\emph{1.47}}} & 91.40\tiny{\emph{1.01}} & 89.76\tiny{\emph{1.25}} & \posLowBox{91.42\tiny{\emph{1.60}}} & 90.10\tiny{\emph{1.90}} & \posMidLowBox{90.96\tiny{\emph{1.73}}} & 89.00\tiny{\emph{2.28}} & \posHighBox{90.57\tiny{\emph{1.74}}} & 89.84\tiny{\emph{1.71}} & \posLowBox{90.29\tiny{\emph{1.68}}} & \posHighBox{91.20\tiny{\emph{1.50}}} \\
& \multirow{-2}{*}{A-UN} & IoU & 83.28\tiny{\emph{2.06}} & 81.91\tiny{\emph{2.15}} & 83.03\tiny{\emph{2.02}} & \posHighBox{85.80\tiny{\emph{1.70}}} & 85.15\tiny{\emph{1.43}} & 82.90\tiny{\emph{1.71}} & \posMidHighBox{86.19\tiny{\emph{1.82}}} & 84.79\tiny{\emph{2.08}} & \posMidLowBox{85.64\tiny{\emph{1.88}}} & 83.74\tiny{\emph{2.32}} & \posMidHighBox{85.06\tiny{\emph{1.94}}} & 83.84\tiny{\emph{1.97}} & \posMidLowBox{84.51\tiny{\emph{1.93}}} & \posHighBox{85.57\tiny{\emph{1.72}}} \\
\cdashline{2-17}
&  & DSC & 90.44\tiny{\emph{1.07}} & 91.02\tiny{\emph{1.46}} & 91.44\tiny{\emph{1.37}} & \posHighBox{92.63\tiny{\emph{0.83}}} & 91.90\tiny{\emph{1.13}} & 91.81\tiny{\emph{1.15}} & \posMidLowBox{92.70\tiny{\emph{0.87}}} & 91.77\tiny{\emph{0.91}} & \negLowBox{91.72\tiny{\emph{1.10}}} & 90.63\tiny{\emph{1.33}} & \posMidHighBox{92.10\tiny{\emph{1.37}}} & 91.52\tiny{\emph{1.44}} & \posMidHighBox{92.55\tiny{\emph{1.06}}} & \posHighBox{92.71\tiny{\emph{0.74}}} \\
\multirow{-6}{*}{COL} & \multirow{-2}{*}{S-UN} & IoU & 83.64\tiny{\emph{1.49}} & 85.17\tiny{\emph{1.68}} & 85.68\tiny{\emph{1.56}} & \posHighBox{86.96\tiny{\emph{1.22}}} & 86.14\tiny{\emph{1.45}} & 86.01\tiny{\emph{1.45}} & \posMidHighBox{87.16\tiny{\emph{1.25}}} & 86.54\tiny{\emph{1.34}} & \posLowBox{85.81\tiny{\emph{1.46}}} & 84.35\tiny{\emph{1.63}} & \posHighBox{86.76\tiny{\emph{1.52}}} & 86.00\tiny{\emph{1.67}} & \posMidHighBox{87.18\tiny{\emph{1.42}}} & \posHighBox{87.01\tiny{\emph{1.15}}} \\

\midrule

&  & DSC & 92.01\tiny{\emph{1.12}} & 92.62\tiny{\emph{1.03}} & 92.67\tiny{\emph{1.00}} & \posMidHighBox{93.80\tiny{\emph{0.63}}} & 93.34\tiny{\emph{0.91}} & 92.98\tiny{\emph{0.71}} & \posLowBox{93.81\tiny{\emph{0.69}}} & 92.53\tiny{\emph{0.96}} & \posMidLowBox{93.48\tiny{\emph{0.76}}} & 92.27\tiny{\emph{1.05}} & \posMidHighBox{93.33\tiny{\emph{0.66}}} & 92.55\tiny{\emph{0.81}} & \posMidHighBox{93.72\tiny{\emph{0.75}}} & \posHighBox{93.64\tiny{\emph{0.71}}} \\
& \multirow{-2}{*}{UN-P} & IoU & 86.80\tiny{\emph{1.28}} & 87.70\tiny{\emph{1.25}} & 87.78\tiny{\emph{1.27}} & \posMidHighBox{88.97\tiny{\emph{0.89}}} & 88.61\tiny{\emph{1.09}} & 87.69\tiny{\emph{1.00}} & \posLowBox{89.09\tiny{\emph{0.95}}} & 87.30\tiny{\emph{1.13}} & \posMidHighBox{88.64\tiny{\emph{1.01}}} & 87.23\tiny{\emph{1.32}} & \posMidLowBox{88.21\tiny{\emph{0.94}}} & 87.15\tiny{\emph{1.11}} & \posHighBox{89.00\tiny{\emph{0.98}}} & \posHighBox{88.85\tiny{\emph{0.98}}} \\
\cdashline{2-17}
&  & DSC & 92.42\tiny{\emph{1.08}} & 92.56\tiny{\emph{1.08}} & 92.74\tiny{\emph{0.73}} & \posMidHighBox{93.61\tiny{\emph{0.70}}} & 93.04\tiny{\emph{1.01}} & 93.12\tiny{\emph{0.82}} & \posMidLowBox{93.98\tiny{\emph{0.68}}} & 93.00\tiny{\emph{1.01}} & \posMidLowBox{93.67\tiny{\emph{0.73}}} & 92.42\tiny{\emph{1.02}} & \posMidHighBox{93.57\tiny{\emph{0.71}}} & 92.56\tiny{\emph{0.88}} & \posMidLowBox{93.36\tiny{\emph{0.75}}} & \posHighBox{94.01\tiny{\emph{0.57}}} \\
& \multirow{-2}{*}{A-UN} & IoU & 87.38\tiny{\emph{1.21}} & 87.70\tiny{\emph{1.28}} & 88.15\tiny{\emph{0.99}} & \posMidHighBox{88.76\tiny{\emph{0.96}}} & 88.34\tiny{\emph{1.19}} & 88.16\tiny{\emph{1.08}} & \posMidHighBox{89.39\tiny{\emph{0.95}}} & 88.26\tiny{\emph{1.19}} & \posMidLowBox{88.87\tiny{\emph{0.94}}} & 87.31\tiny{\emph{1.22}} & \posMidHighBox{88.70\tiny{\emph{0.96}}} & 87.29\tiny{\emph{1.15}} & \posMidHighBox{88.41\tiny{\emph{1.01}}} & \posHighBox{89.23\tiny{\emph{0.82}}} \\
\cdashline{2-17}
&  & DSC & 91.08\tiny{\emph{1.08}} & 92.09\tiny{\emph{0.94}} & 92.34\tiny{\emph{0.95}} & \posMidHighBox{93.38\tiny{\emph{0.80}}} & 92.59\tiny{\emph{1.00}} & 92.84\tiny{\emph{0.94}} & \posMidLowBox{93.49\tiny{\emph{0.85}}} & 93.00\tiny{\emph{0.95}} & \posLowBox{93.47\tiny{\emph{0.83}}} & 92.70\tiny{\emph{0.93}} & \posLowBox{93.03\tiny{\emph{0.92}}} & 91.51\tiny{\emph{1.07}} & \posHighBox{93.20\tiny{\emph{0.88}}} & \posHighBox{93.36\tiny{\emph{0.83}}} \\
\multirow{-6}{*}{CLI} & \multirow{-2}{*}{S-UN} & IoU & 85.17\tiny{\emph{1.29}} & 86.60\tiny{\emph{1.19}} & 87.09\tiny{\emph{1.22}} & \posHighBox{88.53\tiny{\emph{1.03}}} & 87.61\tiny{\emph{1.24}} & 87.90\tiny{\emph{1.19}} & \posMidHighBox{88.83\tiny{\emph{1.09}}} & 88.20\tiny{\emph{1.19}} & \posMidLowBox{88.75\tiny{\emph{1.07}}} & 87.70\tiny{\emph{1.13}} & \posLowBox{88.17\tiny{\emph{1.15}}} & 85.90\tiny{\emph{1.30}} & \posHighBox{88.41\tiny{\emph{1.14}}} & \posHighBox{88.58\tiny{\emph{1.09}}}
\\
\bottomrule
\end{tabular}%
}
\end{table}

%% file: sections/results.tex
\section{Results and Discussion}
\label{sec:results}

\textbf{Datasets:} 
We study 5 medical image \seg datasets from two modalities as labeled data $\dl$: PH2 ($N$=200)~\cite{mendonca2013ph}, Skin Cancer Detection (SCD; $N$=206)~\cite{glaister2013msim}, and DermoFit (DMF; $N$=1,300)~\cite{ballerini2013color} for skin lesion \seg, and CVC-ColonDB (COL; $N$=380)~\cite{bernal2012towards} and CVC-ClinicDB (CLI; $N$=612)~\cite{bernal2015wm} for polyp \seg in colonoscopy. All datasets are split 70:10:20 for train, validation, and test.
As unlabeled data $\du$, we use ISIC2020-Train ($M$=33,126)~\cite{rotemberg2021patient} and Polyp-Box-Seg ($M$=4,070)~\cite{chen2022weakly} for dermatology and colonoscopy respectively: both drawn from different sources than $\dl$. We use 5,000 ISIC2020 images for main experiments (Table~\ref{tab:results}) and the remainder for hyperparameter sensitivity analyses (Table~\ref{tab:ablations}).
All models were trained on Ubuntu 22.04 with Intel Core i9-14900K, 64GB RAM, NVIDIA RTX4090, Python 3.10.19, and PyTorch 2.9.0.


\noindent\textbf{Quality predictor training and evaluation:}
We implement $\mq$ as a ResNet-18 encoder, that takes the channel-wise concatenation of image-mask pairs as input, with a regression head (dropout of 0.15), trained for 150 epochs with AdamW~\cite{loshchilov2018decoupled} (learning rate=3e-4, weight decay=5e-4, batch size 32) with cosine annealing with warm restarts (initial period 10 epochs, doubles after each restart) and early stopping (validation loss; patience 25 epochs) to minimize SmoothL1 loss \rev{(Eqn.~\ref{eqn:l_quality})}~\cite{girshick2015fast}.
We set $p_\mathrm{weak}=0.05$ (Sec.~\ref{subsec:vqm}) and generate $K=50$ degraded masks per sample (Eqn.~\ref{eqn:vqm}).
On test sets, $\mq$ achieves MAE in [0.043, 0.088] and Pearson's correlation coefficient $\rho > 0.92$ across all 5 datasets (Table~\ref{tab:ablations} {$\mathcal{A}$}). 
Zeroing the image input in Eqn.~\ref{eqn:m_q_prediction}, \ie $\mq(x=\bm{0},\tilde{y})$, increases MAE by an average of $0.399 \pm 0.137$ across all 5 datasets (MAE $\in [0,1]$), strongly confirming that $\mq$ leverages image content (contextual grounding), rather than relying on the mask alone.

\input{figs/vis}

\noindent\textbf{SSL \seg training and evaluation:}
Next, we leverage our trained quality predictor to improve \seg using \semisup learning (SSL). We evaluate 3 architectures for $\segmodel$: a purely convolutional model, U-Net++~\cite{zhou2019unet++} (UN-P; 26.08M parameters, 14.14 GFLOPs), a convolutional model with attention gates, Attention U-Net~\cite{oktay2018attention} (A-UN; 24.71M params, 6.16 GFLOPs), and a pure Transformer-based architecture, Swin-U-net~\cite{cao2022swin} (S-UN; 34.27M params, 7.55 GFLOPs), optimizing Dice $+$ cross-entropy loss \rev{(Eqn.~\ref{eqn:seg_loss})}~\cite{isensee2021nnu,taghanaki2019combo} for 200 epochs with AdamW (lr and weight decay set to 1e-4), a cosine annealing scheduler, and early stopping (validation DSC; patience of 30 epochs). We set 
$\lambda_{\mathrm{qar}} = 0.01$ and $\lambda_{\mathrm{qw}} = 0.25$ (Eqn.~\ref{eqn:total_reg},\ref{eqn:total_qw}) and follow a ramp-up schedule during initial training epochs~\cite{tarvainen2017mean}, and report Dice (DSC) and Jaccard (IoU) averaged over 3 runs with different seeds. We compare QAR (Sec.~\ref{subsec:ssl_training} A) against existing popular and widely used SSL paradigms: \pls~\cite{lee2013pseudo} with pixel-level confidence thresholded at 0.9 (PL-T), sample-level confidence-weighted \pls~\cite{sohn2020fixmatch} (PL-C), mean teacher~\cite{tarvainen2017mean} (MT) and its uncertainty-aware extension~\cite{yu2019uncertainty} (UA-MT), interpolation consistency training~\cite{verma2022interpolation} (ICT), \pl-based contrastive learning~\cite{chaitanya2023local} (CL), and cross-pseudo supervision~\cite{chen2021semi} (CPS). 
We do not compare against Zheng et al.~\cite{zheng2020semi}, despite them incorporating quality estimation into SSL, because their code is not available. Nevertheless, we note that their estimator is trained end-to-end with the segmentation model and coupled to a specific self-training pipeline, whereas
ours is the first framework-agnostic approach: an independently trained quality predictor that enhances any \pl-generating method without architectural changes or retraining.
Table~\ref{tab:results} shows that QAR outperforms all competing SSL paradigms across all datasets and models, indicating that \seg quality, even when predicted, may still be a better training signal than model confidence or uncertainty.
It is important to note that
the last two competing methods, CL and especially CPS (as it trains two \seg models in tandem), are computationally more demanding.
Despite no architectural modifications, QAR matches or surpasses state-of-the-art results on these datasets~\cite{du2023mdvit,song2026surmt,dzieniszewska2024improving,liang2026multilevel,perera2025mobileunetr,li2024sutransnet}, many of which employ architectural innovations orthogonal to ours, suggesting further gains from combination.

\input{tables/ablations}

\noindent\textbf{Quality weighting as a drop-in module:}
Next, we show how our quality-weighted \pls can be integrated with any approach that generates \pls $\hat{y}^u_j$ by weighting per-sample losses with $w_j$ (Sec.~\ref{subsec:ssl_training} B). Concretely, for MT~\cite{tarvainen2017mean}: $\mathcal{L}_\mathrm{MT\text{-}QW} = \tfrac{1}{M}\sum_j w_j \lVert \segmodel(x^u_j) - \hat{y}^u_j \rVert^2$; for CPS~\cite{chen2021semi}: $\mathcal{L}_\mathrm{CPS\text{-}QW} = \tfrac{1}{M}\sum_j [ w_{j,2}\, \ell_{\mathrm{seg}}({\segmodel}_1(x^u_j), \hat{y}^u_{j,2}) + w_{j,1}\, \ell_{\mathrm{seg}}({\segmodel}_2(x^u_j), \hat{y}^u_{j,1}) ]$; for ICT~\cite{verma2022interpolation}: $\mathcal{L}_\mathrm{ICT\text{-}QW} = \tfrac{1}{M}\sum_j w_j\, \ell_{\mathrm{seg}}(\segmodel(\tilde{x}^u_j), \tilde{y}^u_j)$; and for CL~\cite{chaitanya2023local}: $\mathcal{L}_\mathrm{CPL\text{-}QW} = \tfrac{1}{M}\sum_j w_j\bigl[\ell_{\mathrm{seg}}(\segmodel(x^u_j), \hat{y}^u_j) + \lambda_c\, \mathcal{L}_\mathrm{contrast}(x^u_j, \hat{y}^u_j)\bigr]$, where \qual weights both the \pl loss and the contrastive loss.
In all cases, $w_j = \mq(x^u_j, \hat{y}^u_j)$ (with $w_{j,k}$ evaluating the \pl from network $k$ in CPS). 
Table~\ref{tab:results} shows that quality-weighted variants ($*$-QW) consistently outperform original versions across all but one setting (DMF + UN-P), confirming the general applicability of our quality-weighting.
\rev{
A scatter plot of the calculated DSC and the predicted quality (Fig.~\ref{fig:vis}), on CLI's test set across all 3 runs of both QAR and PL-QW, shows that the two are strongly correlated. 
Example images (A-D) show two successful $\mq$ predictions for near-perfect (B) and poor (C) \segs, and two failure cases for $\mq$ (A, D).
}

\noindent\textbf{Ablation studies} (Table~\ref{tab:ablations})\textbf{:} Among 5 backbones from different architecture families, ResNet-18 performs the best for $\mq$ ({$\mathcal{B}$}). Incorporating weak-model corruptions (\ie $p_\mathrm{weak} > 0$) improves $\mq$'s performance ({$\mathcal{C}$}), and increasing corrupted masks per sample (\ie $K$) helps up to a saturation point ({$\mathcal{D}$}). $\mq$ trained with weak-model corruptions helps improve \seg model performance for both QAR and PL-QW ({$\mathcal{E}$}). Varying $\lambda_{\mathrm{qar}}$ and $\lambda_{\mathrm{qw}}$ ({$\mathcal{F,G}$}) shows that even suboptimal weights outperform the zero-weight baseline. Finally, increasing unlabeled data ({$\mathcal{H}$}) has minimal impact on final DSC, but substantially accelerates convergence: $\mathrm{e}_{\mathrm{val96}}$ (\ie the number of epochs to reach 96\% validation DSC) decreases considerably with larger $M$.

%% file: figs/vis.tex
\begin{figure*}[!htbp]
    \centering
        \includegraphics[width=\textwidth]{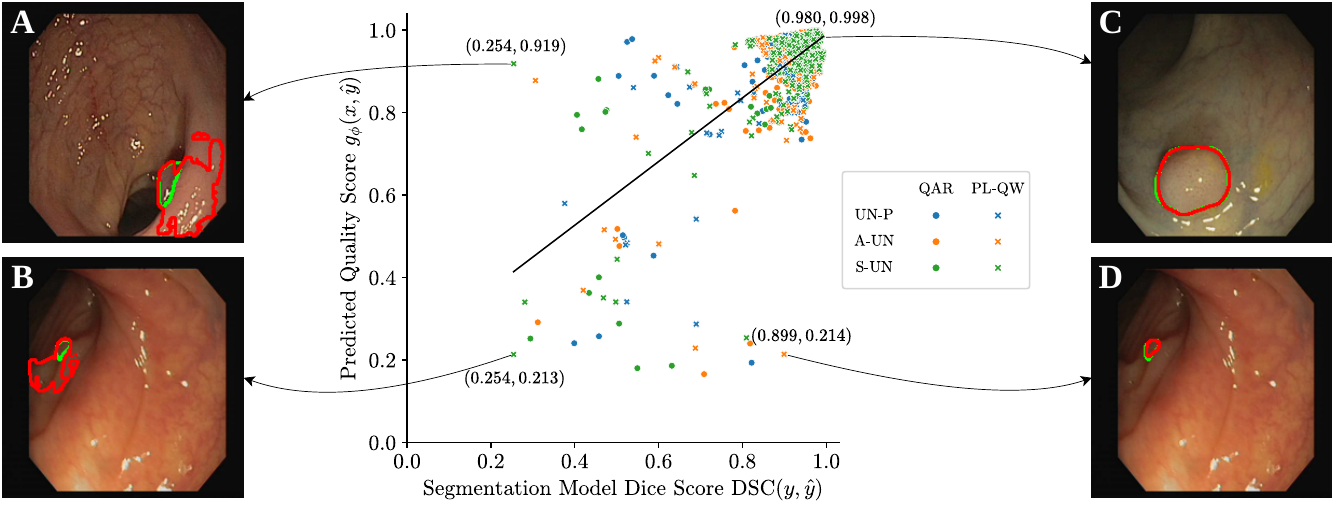}
        \caption{
        \rev{
        Scatter plot of the segmentation predictions' Dice $\mathrm{DSC}(y, \hat{y})$ and the corresponding predicted quality estimates $\mq(x, \hat{y})$ on the test set of CLI dataset, and four representative images (A-D) with the \gt (green) and predicted (red) \segs. We observe a strong, stat. sig., positive linear correlation ($\rho$=0.69; $p$=1e-314).
        }
        }
        \label{fig:vis}
\end{figure*}

%% file: tables/ablations.tex
\begin{table}[!ht]
\centering
\caption{Ablation and hyperparameter sensitivity experiments. The values used for the main experiments table (Table~\ref{tab:results}) are \lightgreyHLBox{highlighted}. Results reported as mean{\tiny \emph{std.err.}}.
Unless specified otherwise, $\dl$ is PH2, $\du$ is ISIC2020-Train, $\mq$ is ResNet-18, $\segmodel$ is Swin-Unet, $p_{\mathrm{weak}}=0.05$, $K=50$, $\lambda_{\mathrm{qar}}=0.01$, $\lambda_{\mathrm{qw}}=0.25$, and $M=5,000$. $\mathrm{e}_{\mathrm{val96}}$ denotes the number of training epochs for the validation DSC to reach 96\%. 
}
\label{tab:ablations}
\resizebox{\textwidth}{!}{%
\setlength{\tabcolsep}{0.5em}
\def\arraystretch{1.05}
\begin{tabular}{@{}lccccc@{}}
\toprule[1pt]
\multicolumn{6}{l}{\greyCellColor\textbf{$\mathcal{A}$. $\mq$'s metrics for all datasets}}                                                                                                                                                    \\
Dataset                                      & PH2                               & SCD                               & DMF                               & COL                               & CLI                               \\\midrule[0.25pt]
MAE / $\rho$                                 & 0.043{\tiny \emph{0.007}} / 0.972 & 0.052{\tiny \emph{0.007}} / 0.973 & 0.060{\tiny \emph{0.004}} / 0.926 & 0.074{\tiny \emph{0.007}} / 0.955 & 0.088{\tiny \emph{0.013}} / 0.927 \\ \midrule[0.75pt]
\multicolumn{6}{l}{\greyCellColor\textbf{$\mathcal{B}$. Varying $\mq$'s backbones (Sec.~\ref{subsec:quality_predictor})}}                                                                                                                                             \\
Backbone                                     & MobileNetV3L                      & EfficientNet-B0                    & \lightgreyHLBox{ResNet-18}                          & DenseNet-121                       & ConvNeXt-T                        \\
Params / FLOPs                               & 4.53M / 0.24G                     & 4.34M / 0.42G                     & 11.31M / 1.87G                    & 7.22M / 2.95G                     & 28.02M / 4.47G                    \\\midrule[0.25pt]
MAE / $\rho$                                 & 0.088{\tiny \emph{0.012}} / 0.849 & 0.054{\tiny \emph{0.009}} / 0.961 & 0.043{\tiny \emph{0.007}} / 0.972 & 0.089{\tiny \emph{0.011}} / 0.837 & 0.055{\tiny \emph{0.008}} / 0.955 \\ \midrule[0.75pt]
\multicolumn{6}{l}{\greyCellColor\textbf{$\mathcal{C}$. Varying $p_{\mathrm{weak}}$ (\ie probability of sampling from weak models' predictions; Sec.~\ref{subsec:vqm})}}                                                                       \\
$p_{\mathrm{weak}}$                          & $0.00$                            & \lightgreyHLBox{$0.05$}                            & $0.10$                            & $0.15$                            & $0.20$                            \\\midrule[0.25pt]
MAE / $\rho$                                 & 0.053{\tiny \emph{0.008}} / 0.960 & 0.043{\tiny \emph{0.007}} / 0.972 & 0.048{\tiny \emph{0.007}} / 0.965 & 0.049{\tiny \emph{0.007}} / 0.953 & 0.051{\tiny \emph{0.009}} / 0.948 \\ \midrule[0.75pt]
\multicolumn{6}{l}{\greyCellColor\textbf{$\mathcal{D}$. Varying $K$ (\ie num. augmented training samples; Eqn.~\ref{eqn:vqm})}}                                                                                                                                                    \\
$K$                                          & $10$                               & $20$                               & \lightgreyHLBox{$50$}                               & $100$                              & $200$                              \\\midrule[0.25pt]
MAE / $\rho$                                 & 0.061{\tiny \emph{0.009}} / 0.937 & 0.048{\tiny \emph{0.006}} / 0.956 & 0.043{\tiny \emph{0.007}} / 0.972 & 0.047{\tiny \emph{0.006}} / 0.969 & 0.050{\tiny \emph{0.005}} / 0.977 \\ \midrule[0.75pt]
\multicolumn{6}{l}{\greyCellColor\textbf{$\mathcal{E}$. Impact of $p_\mathrm{weak} > 0$ on \seg performance}}                                                                                            \\
$p_\mathrm{weak}$                       & $0.00$                            & \cellvrule{\lightgreyHLBox{$0.05$}}                            & $p_\mathrm{weak}$                            & $0.00$                            & \lightgreyHLBox{$0.05$}                            \\\midrule[0.25pt]
QAR: DSC                                          & 94.91{\tiny \emph{0.49}}          & \cellvrule{95.58{\tiny \emph{0.42}}}          & PL-QW: DSC          & 95.10{\tiny \emph{0.46}}          & 95.57{\tiny \emph{0.41}}          \\ \midrule[0.75pt]
\multicolumn{6}{l}{\greyCellColor\textbf{$\mathcal{F}$. Varying $\lambda_\mathrm{qar}$, \ie relative weight of $\lqar$ (Eqn.~\ref{eqn:total_reg})}}                                                                                            \\
$\lambda_\mathrm{qar}$                       & $0.00$                            & \lightgreyHLBox{$0.01$}                            & $0.05$                            & $0.10$                            & $0.20$                            \\\midrule[0.25pt]
DSC                                          & 93.02{\tiny \emph{0.55}}          & 95.58{\tiny \emph{0.42}}          & 95.52{\tiny \emph{0.42}}          & 95.14{\tiny \emph{0.48}}          & 95.03{\tiny \emph{0.49}}          \\ \midrule[0.75pt]
\multicolumn{6}{l}{\greyCellColor\textbf{$\mathcal{G}$. Varying $\lambda_\mathrm{qw}$, \ie relative weight of $\lqw$ (Eqn.~\ref{eqn:total_qw})}}                                                                                               \\
$\lambda_\mathrm{qw}$                        & $0.00$                            & \lightgreyHLBox{$0.25$}                            & $0.50$                            & $0.75$                            & $1.00$                            \\\midrule[0.25pt]
DSC                                          & 93.02{\tiny \emph{0.55}}          & 95.57{\tiny \emph{0.41}}          & 95.25{\tiny \emph{0.44}}          & 95.31{\tiny \emph{0.43}}          & 95.14{\tiny \emph{0.44}}          \\ \midrule[0.75pt]
\multicolumn{6}{l}{\greyCellColor\textbf{$\mathcal{H}$. Varying $M$, \ie number of unlabeled samples (Eqns.~\ref{eqn:l_reg}--\ref{eqn:total_qw})}}                                                                                                                          \\
$M$                                          & $1,000$                           & \lightgreyHLBox{$5,000$}                           & $10,000$                          & $20,000$                          & $30,000$                          \\\midrule[0.25pt]
QAR: DSC / $\mathrm{e}_{\mathrm{val96}}$                                    & 95.48{\tiny \emph{0.43}} / 13         & 95.58{\tiny \emph{0.42}} / 2          & 95.36{\tiny \emph{0.43}} / 1          & 95.50{\tiny \emph{0.43}} / 1          & 95.36{\tiny \emph{0.45}} / 1          \\ \midrule[0.25pt]
PL-QW: DSC / $\mathrm{e}_{\mathrm{val96}}$                                   & 95.48{\tiny \emph{0.43}} / 12          & 95.57{\tiny \emph{0.41}} / 2          & 95.35{\tiny \emph{0.43}} / 2          & 95.48{\tiny \emph{0.43}} / 1          & 95.35{\tiny \emph{0.44}} / 1          \\
\bottomrule
\end{tabular}
}
\end{table}

%% file: sections/conclusion.tex
\section{Conclusion}

We presented a contextually-grounded deep learning-based approach to estimating the \qual of medical image \segs. 
Our \qual predictor is trained on corrupted masks generated using synthetic degradations and weak \seg models' predictions.
We then integrated our \qual predictor into
existing \semisup learning (SSL)-based \seg frameworks through two complementary mechanisms: either as a regularization loss or as a sample reweighting mechanism, without any architectural modifications to the \seg network. Extensive experiments across multiple datasets and model architectures demonstrated
consistent improvements over 
existing SSL paradigms, confirming that learned quality prediction provides an effective training signal for leveraging unlabeled data. Future work could explore extending quality-guided SSL to multi-class \seg, and leveraging \qual predictions for active learning to identify unlabeled samples to be prioritized for expert annotation.